\documentclass{article}

\usepackage[english]{babel}

\usepackage[letterpaper,top=2cm,bottom=2cm,left=3cm,right=3cm,marginparwidth=1.75cm]{geometry}

\usepackage[utf8]{inputenc}
\usepackage[T1]{fontenc}
\usepackage{microtype}

\usepackage{amsmath,amsthm,amsfonts}
\usepackage{graphicx}
\usepackage[colorlinks=true, allcolors=blue]{hyperref}
\usepackage{natbib}
\usepackage{tikz}
\usepackage{pgfplots}

\pgfplotsset{compat=1.10}
\usepgfplotslibrary{fillbetween}
\usetikzlibrary{patterns}

\usepackage{subcaption}
\usepackage{xcolor}

\usepackage[skip=2mm,indent=5mm]{parskip}

\newcommand{\argmin}{\mathop{\rm argmin}}
\newcommand{\argmax}{\mathop{\rm argmax}}

\newtheorem{theorem}{Theorem}
\newtheorem{definition}{Definition}
\newtheorem{proposition}[theorem]{Proposition}
\newtheorem{corollary}[theorem]{Corollary}

\newtheorem{lemma}[theorem]{Lemma}

\newtheorem{claim}{Claim}[section]

\definecolor{DarkGreen}{rgb}{0.1,0.5,0.1}
\definecolor{DarkRed}{rgb}{0.5,0.1,0.1}
\definecolor{DarkBlue}{rgb}{0.1,0.1,0.5}
\hypersetup{
	colorlinks=true,       
	linkcolor=DarkBlue,          
	citecolor=DarkBlue,        
	filecolor=DarkBlue,      
	urlcolor=DarkBlue,          
	pdftitle={},
	pdfauthor={},
}

\newcommand{\cQ}{\mathcal Q}

\newcommand{\cA}{\mathcal A}
\newcommand{\cP}{\mathcal P}

\newcommand{\cZ}{\mathcal Z}
\newcommand{\cX}{\mathcal X}
\newcommand{\cY}{\mathcal Y}

\newcommand{\cH}{\mathcal H}
\DeclareMathOperator*{\E}{\mathbb{E}}

\newcommand{\SR}{S}

\title{Algorithmic Collective Action in Machine Learning\footnote{Authors ordered alphabetically.}}

\usepackage{authblk}
\stepcounter{footnote}
\addtocounter{footnote}{-1}
\author[1]{Moritz Hardt}
\author[2]{Eric Mazumdar}
\author[1]{Celestine Mendler-D\"unner}
\author[3]{Tijana Zrnic}
\affil[1]{Max Planck Institute for Intelligent Systems, T\"ubingen, and T\"ubingen AI Center}
\affil[2]{California Institute of Technology}
\affil[3]{University of California, Berkeley}
\date{}        
\begin{document}

\maketitle

\begin{abstract}
We initiate a principled study of algorithmic collective action on digital platforms that deploy machine learning algorithms. We propose a simple theoretical model of a collective interacting with a firm's learning algorithm. The collective pools the data of participating individuals and executes an algorithmic strategy by instructing participants how to modify their own data to achieve a collective goal.
We investigate the consequences of this model in three fundamental learning-theoretic settings: nonparametric optimal learning, parametric risk minimization, and gradient-based optimization. In each setting, we come up with coordinated algorithmic strategies and characterize natural success criteria as a function of the collective's size. 
Complementing our theory, we conduct systematic experiments on a skill classification task involving tens of thousands of resumes from a gig platform for freelancers. Through more than two thousand model training runs of a BERT-like language model, we see a striking correspondence emerge between our empirical observations and the predictions made by our theory. Taken together, our theory and experiments broadly support the conclusion that algorithmic collectives of exceedingly small fractional size can exert significant control over a platform's learning algorithm.
\end{abstract}

\section{Introduction}

Throughout the gig economy, numerous digital platforms algorithmically profile, control, and discipline workers that offer on-demand services to consumers. Data collection and predictive modeling are critical for a typical platform’s business as machine learning algorithms power ranking, scoring, and classification tasks of various kinds \citep{woodcock2019gig, gray2019ghost, schor2021after}.

Troves of academic scholarship document the emergence and preponderance of precarity in the gig economy. \cite{wood2019good} argue that platform-based algorithmic control can lead to ``low pay, social isolation, working unsocial and irregular hours, overwork, sleep deprivation and exhaustion.'' This is further exacerbated by ``high levels of inter-worker competition with few labor protections and a global oversupply of labor relative to demand.''  In response, there have been numerous attempts by gig workers to organize in an effort to reconfigure working conditions. A growing repertoire of strategies, as vast as it is eclectic, uses both physical and digital means towards this goal. Indeed, workers have shown significant ingenuity in creating platform-specific infrastructure, such as their own mobile apps, to organize the labor side of the platform \citep{chen2018thrown, rahman2021invisible}. Yet, ``the upsurge of worker mobilization should not blind us to the difficulties of organizing such a diverse and spatially dispersed labor force.'' \citep{vallas2020platforms}

Beyond the gig economy, evidence of consumers seeking to influence the algorithms that power a platform's business is abundant. Examples include social media users attempting to suppress the algorithmic upvoting of harmful content by sharing screenshots rather than original posts \citep{twitter}, or individuals creating bots to influence crowd-sourced navigation systems~\citep{sinai14bots}.
The ubiquity of such strategic attempts calls for a principled study of how coordinated groups can wield control over the digital platforms to which they contribute data.

In this work, we study how a collective of individuals can algorithmically strategize against a learning platform. We envision a collective that pools the data of participating individuals and executes an algorithmic strategy by instructing participants how to modify their own data. The firm in turn solves a machine learning problem over the resulting data. The goal of the collective is to redirect the firm's optimization towards a solution that serves the collective. Notably, coordination is a crucial lever. When data are plentiful, a single individual lacks the leverage to unilaterally change the output of a learning algorithm; in contrast, we show that even small collectives can exert substantial influence.

\subsection{Our contribution}

We initiate a principled study of algorithmic collective action in digital platforms that deploy machine learning algorithms. We propose a simple theoretical model of a collective interacting with a firm’s learning algorithm. The size of the collective is specified by a value~$\alpha>0$ that corresponds to the fraction of participating individuals in a population drawn from a base distribution~$\cP_0.$
The firm observes the mixture distribution
\[\cP=\alpha \cP^* + (1-\alpha) \cP_0,\]
where $\cP^*$ depends on the strategy of the collective, and runs a learning algorithm $\cA$ on $\cP$.

We investigate the consequences of this model in three fundamental learning-theoretic settings, supported by a systematic empirical evaluation on a real resume classification task from a gig platform. In each setting, we come up with algorithmic collective strategies and characterize different success criteria as a function of the collective’s size~$\alpha$. We exhibit \emph{critical thresholds} for the value of~$\alpha$ at which the collective succeeds. Our theory and experiments support the conclusion that collectives of vanishing fractional size can exert significant control over the firm’s learning algorithm.

\paragraph{Classification.}
In line with economic theory, we start with the case of an optimal firm that has full knowledge of the distribution $\cP$. The firm chooses the Bayes optimal predictor~$f=\cA(\cP)$ on the distribution~$\cP$.  In the context of classification, 
a natural class of objectives for the collective is to correlate a signal $g(\cdot)$ with a target label $y^*$: an individual described by data point $(x,y)$ succeeds if $f(g(x))=y^*$ at test time.
For this goal, a natural strategy is to perturb each participating data point $(x,y)$ by applying the signal $g(\cdot)$ to $x$ and switching the label from $y$ to $y^*$ at training time. That is, $\cP^*$ is the distribution of $(g(x), y^*)$ for a random draw of a labeled data point~$(x, y)$ from~$\cP_0$. We prove that a collective of vanishing fractional size succeeds with high probability by implementing this strategy, provided that the signal $g(x)$ is unlikely to be encountered in the base distribution~$\cP_0$. The success probability is maximized for an optimal classifier and deteriorates gracefully with the suboptimality of the classifier.

In practice, the signal~$g(x)$ may correspond to adding a hidden watermark in image and video content, or subtle syntactic changes in text. It is reasonable to assume that individuals are indifferent to such inconsequential changes to their features. In fact, conventional wisdom in machine learning has it that such hidden signals are easy to come by in practice. The ability to change the label of an instance, however, is a more strenuous requirement. We therefore propose a variant of the strategy where a participating individual does not need to modify the label of their data point. We show that this strategy still succeeds, while quantifying precisely how it diminishes the collective’s success rate as a function of its size and a key parameter of $\cP_0$.

We provide analogous results when the collective's goal is for the firm's predictions to ignore some subset of the feature information. Given a map $g(x)$, the collective succeeds if $f(x)=f(g(x))$. Here, $g(x)$ is a summary of $x$ that, for example, removes private or sensitive information in~$x$.

\paragraph{Experimental evaluation.}
We conduct extensive experiments on a dataset of almost thirty thousand resumes from a gig platform for freelancers. The machine learning task is to tag resumes with a set of ten skills related to work in the IT sector. Through more than two thousand model training runs involving a BERT-like language model, we investigate the predictions made by our theory. What emerges is a striking correspondence between our empirical findings and the theory. The ability to modify both resume and label leads to a near-perfect success rate at collective sizes of only a fraction of a percent of the population, corresponding to fewer than one hundred modified resumes. The label-free strategy still succeeds reliably, albeit at a higher threshold, corresponding to a few hundred resumes. The more well-trained the model is, the lower the threshold for the collective's success. The placement pattern of the signal in the resume is largely irrelevant, so long as the token we plant is unique within the corpus of resumes.

Our theory predicts that collectives, in a certain precise sense, must compete with the strongest alternative signal in the data. The weaker the alternative signals, the lower the threshold for success. To confirm this hypothesis experimentally, we randomize the labels of a random fraction of the data. Confirming our theory, we observe that increasing the fraction of randomized labels, and hence diminishing the strength of alternative signals, indeed lowers the threshold for success of the collective. This observation suggests a blessing of dimensionality: if the data contains many weak signals, as high-dimensional data tends to, algorithmic collectives are especially powerful.

\paragraph{Risk minimization and gradient descent.}
Generalizing beyond classification, we consider the power of collectives in convex risk minimization and gradient-based learning with possibly nonconvex objectives. In the first case, we show that collectives can force the firm to deploy any model with small suboptimality on $\cP_0$ of the collective's choosing. 
In the second case, we show that given a more powerful interaction model the collective can influence the firm to deploy any desired model even under nonconvexity, as long as the path from the initial model, computed on $\cP_0$, to the desired model does not traverse large gradients when evaluated on $\cP_0$. Moreover, despite the nonconvexity, convergence to the target is achieved at a convex rate. In both problem settings, the analyzed collective strategies rely on exerting influence on the gradients computed by the firm.

\subsection{Related work}

Our approach to algorithmic collective action is decidedly not adversarial. Instead, the strategic manipulations arise through a misalignment of the firm's and the individuals' objectives. Individuals legitimately optimize their utility through data sharing and coordination. Yet, at a technical level our results relate to topics studied under the umbrella of adversarial machine learning. Most closely related is the line of work on \emph{data poisoning attacks} that seeks to understand how data points can be adversarially ``poisoned'' to degrade the performance of a predictive model at test time. 
We refer to recent surveys for an overview of data poisoning attacks~\citep{zhiyi22survey}, and backdoor attacks more specifically~\citep{guo22backdoor}. 
While the literature on poisoning attacks focuses predominantly on diminishing the performance of the learning algorithm, documented empirical successes \citep{cherepanova2021lowkey,geiping2021witches} hint at the impact that algorithmic collective action can have on deep learning models.
Despite the increasing number of studies on backdoor attacks and defenses, theoretical work explaining how underlying factors affect the success of backdoor attacks has been limited~\citep{grosse22demystify}. 
We point out two recent works that study the relationship between the number of manipulated data points and the success probability of the attack. 
\cite{manoj2021excess} show that a fixed number of points is sufficient for backdoor attacks to succeed in binary classification problems, provided that the memorization capacity of the model is sufficiently large.
\cite{cina21learningcurves} empirically investigate backdoor learning curves across many image recognition tasks and they observe curves with diminishing returns, similar in shape to those in our experiments.
Our analysis of Bayes optimal classification provides a new, complementary theoretical perspective and sheds light on the effectiveness of practical data manipulation strategies in a way that is surprisingly predictive of our empirical observations and previous empirical results in adversarial machine learning.

Our analysis of risk minimization is reminiscent of model-targeted attacks~\citep{suya21model,farhadkhani22grad}, which aim to bias the learner towards a target model. Our gradient-control strategy resembles the counter-gradient attack by~\citet{farhadkhani22grad}.
While the insights from these prior works  are valuable to inform the feasibility of collective action in convex risk minimization, our work differs from these existing studies in its focus on the role of collective size and the analysis of nonconvex losses.
Only a handful of works in the adversarial machine learning literature have questioned the institution-centric perspective of the community and discussed the political dimension of adversarial machine learning~\citep{Albert2020PoliticsOA,vincent21leverage,kendra21sop}. In this context, a recent line of work considers the socially beneficial use of adversarial learning techniques, e.g.,~\citep{delobelle21ethical, shan20fawkes, abebe2022adversarial, kulynych20pot,li2022untargeted}. 


Our work can also be viewed as a conceptual reversal of \emph{strategic classification}~\citep{hardt16strategic}. In strategic classification, a firm anticipates the optimal response of a strategic individual to a decision rule. Instead, we consider individuals that strategically anticipate the optimizing behavior of the firm, something recently considered by~\cite{zrinc21follow}. Furthermore, our work is conceptually different from strategic classification in its focus on the role and objectives of workers and consumers on online platforms rather than the firm.  Another crucial departure from the set of problems considered in strategic classification is our emphasis on collective rather than individual strategies.

The idea of collective action on digital platforms has also been previously studied. \cite{elliot21recourse} show how algorithmic recourse can be improved through coordination.  \citet{vincent2019data} examine the effectiveness of \emph{data strikes}. Extending this work to the notion of \emph{data leverage}, \citet{vincent21leverage} describe various ways of ``reducing, stopping, redirecting, or otherwise manipulating data contributions'' for different purposes. See also \citet{vincent21CDC}. Our work provides a theoretical framework for understanding the effectiveness of such strategies, as well as studying more complex algorithmic strategies that collectives may deploy. 

Appendix~\ref{sec:extended-related} continues our discussion of related work.

\section{Problem formulation}
We study the strategic interaction of a firm operating a predictive system with a population of individuals. We assume
that the firm deploys a learning algorithm~$\cA$ that operates on data points in a universe~$\cZ=\cX\times\cY$. Each individual corresponds to a single data point $z\in\cZ$, typically a feature--label pair. We model the population of individual participants as a distribution $\cP_0$ over $\cZ$. 

We say that a fraction $\alpha>0$ of the individuals form a \emph{collective} in order to strategically respond to the firm's learning behavior.
The collective agrees on a potentially randomized strategy $h:\cZ\rightarrow\cZ$ from a space of available strategies $\cH$. The possible strategies $\cH$ capture feasible changes to the data. 
For example, content creators on a video streaming platform may be indifferent between videos that differ only in a hidden watermark not visible to human viewers. Freelancers may be indifferent between two resumes that differ only in inconsequential syntactic details.

The firm therefore observes a mixture distribution
\begin{equation*}
\cP = \alpha \cP^* + (1-\alpha) \cP_0,\,
\end{equation*}
where we use $\cP^*$ to denote the distribution of $h(z), z\sim \cP_0$.

The collective strives to choose a strategy $h$ so as to maximize a measure of success over the solution~$f=\cA(\cP)$ chosen by the firm. Here, $f$ is a mapping from features to labels, $f:\cX\rightarrow\cY$. Given a strategy, we use $S(\alpha)$ to denote the level of success achieved by a collective of size $\alpha$. The central question we study is how the success $S(\alpha)$ grows as a function of collective size~$\alpha,$ and how large $\alpha$ needs to be in order to achieve a target success level.

\begin{definition}[Critical mass] The critical mass for a target success level $S^*$ is defined as the smallest $\alpha$ for which there exists a strategy such that $S(\alpha)\geq S^*$.
\end{definition}
Note that, although motivated from the perspective of labor, our formal model can also serve as a basis for studying collective action on the consumer side of digital platforms.
Before presenting our results we briefly discuss why we focus on collective strategies in this paper.

\paragraph{Why collective action?} By engaging in collective action, individuals can exert influence on the learning algorithm that they could not achieve by acting selfishly. In large-population settings such as online platforms, an individual contributes an infinitesimal fraction of the data used by the learning algorithm. Thus, under reasonable manipulation constraints, individual behavior is largely powerless in systematically changing the deployed model. Instead, individuals are limited to simple adversarial attacks or individual strategies that do not have lasting effects on the learning outcome. By coordinating individuals, however, collectives can wield enough power to steer learning algorithms towards desired goals. In subsequent sections we show that collectives can often do so while only representing a small fraction of the training data.

\section{Collective action in classification}

We start with classification under the assumption that the firm chooses an approximately optimal classifier on the data distribution~$\cP$.

\begin{definition}[$\epsilon$-suboptimal classifier]
\label{def:clf-opt}
{
A classifier~$f\colon\cX\to\cY$ is $\epsilon$-suboptimal on a set $\cX'\subseteq\cX$ under the distribution~$\cP$  if there exists a $\cP'$ with $\mathrm{TV}(\cP_{Y|X=x},\cP'_{Y|X=x})\leq \epsilon$  such that for all $x\in \cX'$
\begin{equation*}
\label{eq:bayes-opt}
f(x) = \argmax_{y\in\cY} \cP'(y|x)\,.
\end{equation*}
}
\end{definition}
\noindent
If we say that a classifier is $\epsilon$-suboptimal without specifying the set $\cX'$, then we implicitly assume that it is $\epsilon$-suboptimal for every $x\in\cX$. Note that a $0$-suboptimal classifier is the Bayes optimal classifier with respect to the zero--one loss. In turn, it is not hard to see that every classifier is $0.5$-suboptimal. Thus, without loss of generality we assume $\epsilon\leq 0.5$ in the following.

Under the above assumption, we focus on two general goals for the collective: \emph{planting a signal} and \emph{erasing a signal}.

\subsection{Planting a signal}
\label{sec:trigger}

Assume the collective wants the classifier to learn an association between an altered version of the features $g(x)$ and a chosen target class $y^*$. Formally, given a transformation $g:\cX\rightarrow\cX$, the collective wants to maximize the following measure of success:
\[
S(\alpha)=\Pr_{x\sim \cP_0} \left\{ f(g(x)) = y^*\right\}.
\]
We call this objective ``planting a signal'' and $\cX^*=\{g(x)\colon x\in\cX\}$ the signal set. For example, $g(x)$ could be instance $x$ with an inconsequential trigger (such as a video with an imperceptible watermark or a resume with a unique formatting) and $y^*$ could be a label indicating that the instance is of high quality (e.g., a high-quality video or a highly qualified individual). 
As another example, the collective may have an altruistic goal to help individuals in a vulnerable subpopulation $\cX_0\subseteq\cX$ achieve a desired outcome $y^*$. In this case,
$g(x)$ could be a mapping from $x$ to a randomly chosen instance in~$\cX_0$.

We provide natural strategies for planting a signal and characterize their success as a function of $\alpha$. The key parameter that we identify as driving success is the \emph{uniqueness} of the signal.

\begin{definition}[$\xi$-unique signal] We say that a signal is $\xi$-unique if it satisfies $\cP_0(\cX^*)\leq \xi$.
\end{definition}

In addition, success naturally depends on how suboptimal $y^*$ is on the signal set under the base distribution. To formalize this dependence, we define the suboptimality gap of $y^*$ as
$$\Delta = \max_{x\in\cX^*}\left(\max_{y\in\cY}\,\cP_0(y|x) - \cP_0(y^*|x)\right)\,.$$
We consider two possibilities for the space of available strategies $\cH$. First, we assume that the individuals can modify both features and labels. We call the resulting strategies \emph{feature--label} strategies. Modifying features by, say, planting a trigger often comes at virtually no cost. Changing the label, however, may be hard, costly, or even infeasible. This is why we also consider \emph{feature-only} strategies; such strategies only allow changes to features.

\paragraph{Feature--label signal strategy.} We define the feature--label signal strategy as 
\begin{equation}
\label{eq:trigger-label}
h(x,y)=(g(x), y^*)\,.
\end{equation}
The result below quantifies the success of this strategy in terms of the collective size and the uniqueness of the signal.

\begin{theorem}
\label{thm:trigger-label}
Consider the feature--label signal strategy and suppose that the signal is $\xi$-unique. Then, the success against a classifier that is  $\epsilon$-suboptimal on the signal set $\cX^*$ is lower bounded by
{
\[\SR(\alpha) \ge 1 - \frac{1-\alpha}{\alpha} \cdot \frac{(1-\epsilon)\Delta + \epsilon}{1-2\epsilon} \cdot \xi \,.\]
}
\end{theorem}

Rearranging the terms, we obtain an upper bound on the critical mass given a desired success probability (e.g., $90\%$).

\begin{corollary}
Suppose the signal is $\xi$-unique. Then, the critical mass for achieving success $\SR^*\in(0,1)$ with feature--label strategies against a classifier that is $\epsilon$-suboptimal on the signal set $\cX^*$ is bounded~by
{
\begin{equation}
\label{eq:feature-label-alpha}
\alpha^* \leq \frac{\frac{(1-\epsilon)\Delta + \epsilon}{1-2\epsilon}\cdot  \xi}{1 -\SR^* + \frac{(1-\epsilon)\Delta + \epsilon}{1-2\epsilon} \cdot \xi}.
\end{equation}
}
\end{corollary}

Therefore, in order to achieve success it suffices to have a collective size proportional to the uniqueness of the signal and the suboptimality of $y^*$ on the signal set, as long as these parameters are sufficiently small relative to the target error rate $1-S^*$. This suggests that planting signals that are exceedingly ``rare'' under the base distribution can be done successfully by small collectives--- a property of feature--label strategies that we empirically validate in Section~\ref{sec:exps}.

\begin{figure}
\begin{center}
\includegraphics[width=0.35\columnwidth]{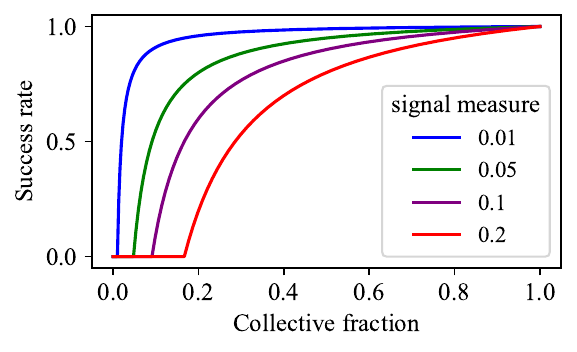}
\includegraphics[width=0.35\columnwidth]{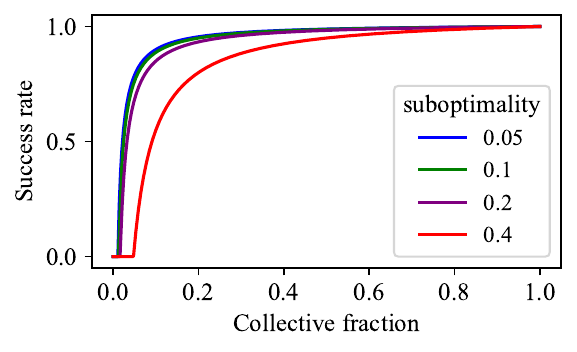}
\end{center}
\caption{Illustration of the success rate predicted by Theorem~\ref{thm:trigger-label}. In the first we fix $\epsilon=0$ and vary $\xi$, and in the second we fix $\xi$ and vary the classifier's suboptimality, $\epsilon$. We upper bound the suboptimality gap as $\Delta\leq 1$.}
\end{figure}

In the next result we consider feature-only strategies. 
An impediment to the success of such strategies is the situation where there is overwhelmingly strong signal about a label $y\neq y^*$ in the base distribution and hence little label uncertainty. This is the reason why we make one additional assumption that there exists a number $p>0$ such that $\cP_0(y^*|x)\ge p$ for all $x\in\cX$.

\paragraph{Feature-only signal strategy.} We define the feature-only signal strategy as
\begin{equation}
\label{eq:trigger-only}
h(x,y)=
\begin{cases}
(g(x), y^*), & \text{ if } y=y^*,\\
(x, y),      & \text{otherwise}.
\end{cases}
\end{equation}
This strategy achieves a similar success rate as the feature--label strategy, but the success diminishes with the constant~$p$. 
\begin{theorem}
\label{thm:trigger-only}
Consider the feature-only signal strategy and suppose that the signal is $\xi$-unique. Further, suppose there exists $p>0$ such that $\cP_0(y^*|x)\ge p,\forall x\in\cX$. Then, the success against a classifier that is $\epsilon$-suboptimal on the signal set $\cX^*$ is lower bounded by
{
\[S(\alpha) >1- \frac{1-(1-\epsilon)p-\epsilon\alpha}{(1-\epsilon)p\alpha-\epsilon\alpha}\; \xi.\]
}
\end{theorem}

The critical mass for achieving a target success probability is thus bounded as follows.

 \begin{corollary}
Suppose the signal is $\xi$-unique. Then, the critical mass for achieving success $\SR^*\in(0,1)$ with feature-only strategies against a classifier $\epsilon$-suboptimal on the signal set $\cX^*$ is bounded~by
\begin{equation}
\label{eq:feature-only-alpha}
\alpha^* \leq \frac{(1-(1-\epsilon)p)\xi}{(1- S^*)\left((1-\epsilon)p-\epsilon\right)+\epsilon\xi}.
\end{equation}
\end{corollary}

Whenever the positivity constant $p$ is smaller than $0.5$, the critical mass \eqref{eq:feature-only-alpha} that guarantees success of feature-only strategies is at least as large as the critical mass \eqref{eq:feature-label-alpha} for feature--label strategies, as expected.
{{Similar to the feature--label strategy, the success depends on the optimality of the classifier on $\cX^*$, offering another lever for the collective to optimize their strategy.}}

The positivity constant $p>0$ may be excessively small over the entire data universe. A standard fix to this problem is to restrict $\cP_0$ to a subset where the constant is larger, and pay a penalty for the amount of truncation in the bound. For example, if there exists $R\subseteq\cX$ such that $\cP_0(R)\geq 99\%$, but the positivity constant over $R$ is much larger than $p$, then one can obtain a more powerful version of Theorem~\ref{thm:trigger-only}. 

\subsection{Erasing a signal}

Next, we assume the collective wants the classifier to be invariant under a transformation $g:\cX\rightarrow\cX$ of the features. In particular, the success is measured with respect to:
\[\SR(\alpha)= \Pr_{x\sim\cP_0} \{f(x)=f(g(x))\}.\]
In other words, the collective wants the classifier to output the same predictions for all $x$ and $x'$ that have $g(x)=g(x')$. The map $g$ can be thought of as a summary of $x$ that removes some feature information. We call this objective ``erasing a signal.'' For example, if the collective wants the deployed model to be insensitive to the value of a particular feature $j^*$, then it can use $g(x) = x'$ where $x'_j = x_j$ for $j\neq j^*$ and $x'_{j^*} = 0$. The feature $j^*$ could be the length of a video that content creators do not want to affect the ranking of the content, or it could be a sensitive demographic feature that a collective wants to be independent of the predicted label.

\paragraph{Erasure strategy.} We define the erasure strategy as
\[h(x,y)=\left(x,\argmax_{y\in\cY} \cP_0(y|g(x)) \right).\]
As before, the success of the strategy depends on problem-dependent quantities. In this case, the quantity of interest is  the sensitivity of the labels to the erased signal. We capture this sensitivity in the parameter $\tau$, defined as
\[\tau = \E_{x\sim\cP_0} \max_{y\in\cY}\left|\cP_0(y|x)-\cP_0(y|g(x))\right|.\]
Intuitively, $\tau$ is small if observing the whole feature vector $x$, instead of just the summary $g(x)$, reveals little additional information about the label. 

\begin{theorem}\label{thm:dissocation}
Consider the erasure strategy. Then, the success against an $\epsilon$-suboptimal classifier is lower bounded by
{
\[\SR(\alpha)\geq 1 - \frac{2(1-\alpha)}{\alpha} \cdot \tau - \frac{\epsilon}{(1-\epsilon) \alpha}.\]
}
\end{theorem}
We rearrange the terms and derive a bound on the critical mass that guarantees a signal can be erased with a desired probability.

\begin{corollary}
The critical mass for achieving success $S^*\in~(0,1)$ against an $\epsilon$-suboptimal classifier is bounded by
{
    \[\alpha^*\leq \frac{2\tau + \frac{\epsilon}{1-\epsilon}}{1-S^* + 2\tau}.\]
}
\end{corollary}

The less sensitive the labels to the erased information, the smaller the collective needed to successfully enforce a decision rule independent of the protected information. 

In contrast to the strategies in Section~\ref{sec:trigger}, the erasure strategy requires knowledge of statistics about $\cP_0$. This highlights an important benefit of collective action: information sharing. Information about the base distribution is typically difficult to obtain for individual platform users. However, a collective can pool their feature--label information to estimate properties of the distribution from samples; the larger the collective, the better the estimate and consequently the more effective the strategy.

\section{Collective action in risk minimization}
\label{sec:risk-min}

We next study the effect of collective size when the learner is solving parametric risk minimization. Here the firm is choosing a model from a parameterized set $\{f_\theta\}_{\theta\in\Theta}$. We will use $\cA(\cP)$ to denote an element in $\Theta$ that determines the model chosen by the firm.
We begin by studying convex risk minimizers. Then, motivated by nonconvex settings, we look at gradient-descent learners without imposing any convexity assumptions on the objective.
Our main working assumption will be that of a risk-minimizing firm.

\begin{definition}[Risk minimizer]
Fix a loss function $\ell$. The firm is a risk minimizer if under distribution $\cP$ it determines the parameter of the model $f_\theta$ according to
\[\theta = \argmin_{\theta'\in\Theta}\; \E_{z\sim \cP} \ell(\theta';z).\]
\end{definition}

\subsection{Convex risk minimization}

To begin, we assume that $\ell(\theta;z)$ is a convex function of $\theta$, and that the collective's goal is to move the model from $\theta_0$---the optimal model under the base distribution $\cP_0$---to a target model $\theta^*$. To that end, for a given target model $\theta^*\in\Theta$, we measure success in terms of
\[S(\alpha)= - \|\theta-\theta^*\|.\]
Here, $\|\cdot\|$ can be any norm (as long as it is kept fixed in the rest of the section). 
In line with first-order optimality conditions for convex optimization, the influence of the collective on the learning outcome depends on the collective's ability to influence the average gradient of $\ell$.
To simplify notation, let $g_{\cP}(\theta') = \E_{z\sim \cP}\nabla \ell(\theta';z)$
denote the expected gradient of the loss over distribution $\cP$ measured at a point $\theta'\in\Theta$.

\paragraph{Gradient-neutralizing strategy.} Define the gradient-neutralizing strategy as follows. Find a \emph{gradient-neutralizing} distribution $\cP'$ for $\theta^*$, meaning $\angle(g_{\cP'}(\theta^*), - g_{\cP_0}(\theta^*))=0$. Then, draw $z'\sim\cP'$ and let
\[
h(z) =\begin{cases}
z', &\text{with probability }\min\left(1,\frac{1}{\alpha} \frac{\|g_{\cP_0}(\theta^*)\|}{\|g_{\cP'}(\theta^*)\| + \|g_{\cP_0}(\theta^*)\|} \right),\\
z, & \text{else}.
\end{cases}
\]

For example, in generalized linear models (GLMs) gradients are given by $\nabla \ell(\theta;(x,y)) = x(\mu_\theta(x) - y)$, where $\mu_\theta(\cdot)$ is a mean predictor (see, e.g., Chapter 3 in \citep{efron2022exponential}). Therefore, one can obtain a gradient-neutralizing distribution by simply letting $h(x,y) = (x',y')$, where $x' = - g_{\cP_0}(\theta^*)$ and $y'$ is any value less than $\mu_\theta(x')$. Alternatively, if the collective is restricted to feature-only strategies, they can set $x' = - g_{\cP_0}(\theta^*)$ only if $y< \mu_\theta(x')$, and $x'=0$ otherwise. As long as the label distribution has sufficiently large support under $\cP_0$, in particular $y< \mu_\theta(- g_{\cP_0}(\theta^*))$ with nonzero probability, this strategy likewise results in a gradient-neutralizing distribution.

\begin{theorem}
\label{thm:convex}
Suppose there exists a gradient-neutralizing distribution $\cP'$ for $\theta^*$. Then, if the loss is $\mu$-strongly convex, the success of the gradient-neutralizing strategy is bounded~by
\[S(\alpha)\geq \frac{1}\mu\min\left(\alpha \|g_{\cP'}(\theta^*)\| - (1-\alpha)\|g_{\cP_0}(\theta^*)\|,0\right).\]
\end{theorem}

The natural target success for the collective is for $\theta^*$ to be reached exactly; this is achieved when $S(\alpha)=0$.
\begin{corollary}
\label{cor:convex}
Suppose there exists a gradient-neutralizing distribution $\cP'$ for $\theta^*$. Then, for any $\mu\geq 0$ the critical mass for achieving target success $S(\alpha)=0$ is bounded by
\begin{equation}
\label{eq:alpha_convex}
\alpha^*\leq \frac{\|g_{\cP_0}(\theta^*)\|}{\|g_{\cP'}(\theta^*)\| + \|g_{\cP_0}(\theta^*)\| }.
\end{equation}
\end{corollary}
Even if $\ell$ is only strictly convex  ($\mu\rightarrow 0$), the collective can reach $\theta^*$ with $\alpha^*$ as in~\eqref{eq:alpha_convex}. Note that this corollary reveals an intuitive relationship between $\alpha^*$ and $g_{\cP_0}(\theta^*)$ in the convex regime: target models $\theta^*$ that look more optimal to the learner under the base distribution are easier to achieve.

If the collective has a utility function $u(\theta')$ that specifies the participants' preferences over different models $\theta'$, a natural way to define success is via a desired gain in utility: 
\[S(\alpha) = u(\theta) - u(\theta_0),\] 
where $\theta_0 = \cA(\cP_0)$. Corollary \ref{cor:convex} implies a bound on the critical mass for this measure of success, for all convex utilities (for example, linear utilities of the form $u(\theta) = \theta^\top v$, for some $v$).

\begin{proposition}
\label{prop:collective-utility}
Suppose that $u(\theta')$ is convex. Further, assume $\ell$ is $\beta$-smooth and that $\|\cdot\|$ is the $\ell_2$-norm. Then, the critical mass for achieving $u(\theta) - u(\theta_0)\geq U$ is bounded~by
$$\alpha^* \leq \frac{\beta\cdot U}{g_{\mathrm{lb}} \cdot\|\nabla u(\theta_0)\| + \beta\cdot U},$$
where $g_{\mathrm{lb}} = \min\{ \|g_{\cP'}(\theta')\|: \|\theta'-\theta_0\|\leq U/\|\nabla u(\theta_0)\| \}$ and $\cP'$ is a gradient-neutralizing distribution for model $\theta'$.
\end{proposition}

As a result, the critical mass required to achieve a utility gain of $U$ decreases as the gradient of the utility at $\theta_0$ grows. Intuitively, large $\|\nabla u(\theta_0)\|$ means that small changes to $\theta_0$ can lead to large improvements for the collective.

\subsection{Gradient-based learning}

So far we have assumed that exact optimization is computationally feasible; with nonconvex objectives, this behavior is hardly realistic. A common approach to risk minimization for general, possibly nonconvex learning problems is to run gradient descent.

Formally, at each step $t$ we assume the learner observes the current data distribution $ \cP_t$, computes the average gradient at the current iterate, and updates the model by taking a gradient step:
\[\theta_{t+1}=\theta_t -\eta \cdot g_{\cP_t}(\theta_t),\]
where $\eta>0$ is a fixed step size.
Given a target model $\theta^*$, we define the success of the strategy after $t$ steps as
\[S_t(\alpha)=-\|\theta_t-\theta^*\|.\]
Given the online nature of the learner's updates, we assume that the collective can \emph{adaptively} interact with the learner; that is, they can choose $\cP_t^*$ at every step $t$. This is a typical interaction model in federated learning~\citep{mcmahan17federated}. In the following we show that this additional leverage enabled by this adaptivity allows the collective to implement a refined strategy that controls the outcome of learning even in nonconvex settings.

\paragraph{Gradient-control strategy.} We define the gradient-control strategy at $\theta$ as follows. Given the observed model $\theta$ and a target $\theta^*$, the collective finds a \emph{gradient-redirecting distribution} $\cP'$ for $\theta$, meaning
$g_{\cP'}(\theta) = -\frac{1-\alpha} \alpha g_{\cP_0}(\theta) + \xi(\theta-\theta^*)$, for some $\xi\in(0,\frac{1}{\alpha\eta})$. Then, draw $z'\sim\cP'$ and set 
\[h(z)=z'.\]

The gradient-control strategy is easiest to implement when $\frac{1-\alpha} \alpha \|g_{\cP_0}(\theta)\|$ is small; then, it is reasonable to expect to find $\cP'$ that neutralizes the small effect. If the collective size $\alpha$ is small or the gradients $\|g_{\cP_0}(\theta)\|$ are large, it becomes increasingly difficult to find a gradient-redirecting distribution.

If the collective can supply gradients directly rather than implicitly through data points (as in the Byzantine learning setting~\citep{blanchard17byzantine}), there is no need for a gradient-redirecting distribution and the gradient-control strategy is implemented by supplying gradients so that the average gradient of the collective $\bar g$ satisfies $\bar g = -\frac{1-\alpha} \alpha g_{\cP_0}(\theta) + \xi(\theta-\theta^*)$.

\begin{theorem}
\label{thm:GD}
Assume the collective can implement the gradient-control strategy at all $\lambda \theta_0 + (1-\lambda) \theta^*, \lambda\in[0,1]$. Then, there exists a $C(\alpha)>0$ such that the success of the gradient-control strategy after $T$ steps is lower bounded by 
\[S_T(\alpha)\geq -\left(1- \eta C(\alpha)\right)^T\cdot \|\theta_0-\theta^*\|.\]
\end{theorem}
The above result implies that the collective can reach any model $\theta^*$ as long as there exists a path from $\theta_0$ to $\theta^*$ that only traverses small gradients on the initial distribution $\cP_0$.

\begin{figure*}[t!]
\begin{center}
\includegraphics[width=0.9\textwidth]{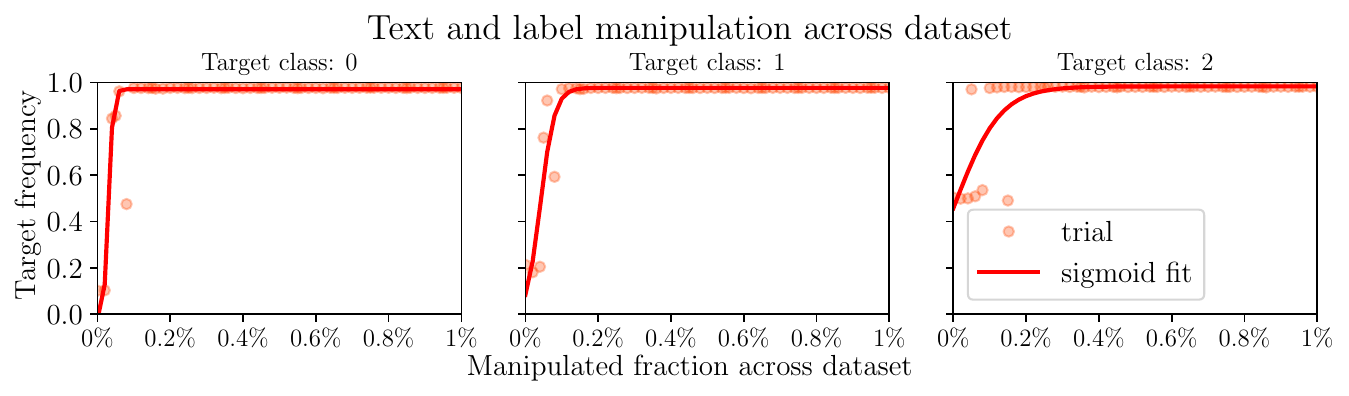}
\includegraphics[width=0.9\textwidth]{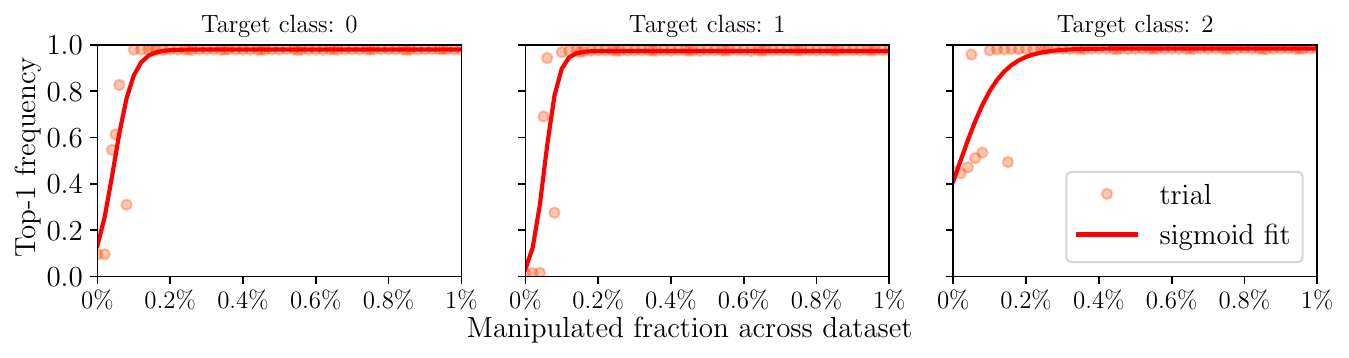}
\end{center}
\caption{Success rate of Strategy 1 as the collective size varies. Each dot represents one model training run. The solid line is a best-fit sigmoid function with two shape parameters and one offset term.}
\label{fig:trigger-label}
\end{figure*}

\section{Experiments}
\label{sec:exps}

We report on experimental findings from over $2000$ model training runs involving a BERT-like text transformer model on a resume classification task. The resume dataset consists of nearly $30000$ resumes of freelancers on a major gig labor platform, introduced by \cite{jiechieu2021skills}. The task is a multiclass, multilabel classification problem where the goal is to predict a set of up to ten skills from the software and IT sector based on the resume. 

The collective controls a random fraction of the training data within the dataset. Its goal is to plant a signal, that is, steer the model's predictions on transformed data points $g(x)$ toward a desired target label $y^*$. We evaluate the effectiveness of two simple strategies, which are instantiations of the feature--label and feature-only strategies from Section \ref{sec:trigger}.

\paragraph{Strategy 1 (Text and label manipulation across dataset).} 
The collective plants a special token in the resume text and changes the label to the target class. 
This strategy closely mirrors the feature-label signal strategy in~\eqref{eq:trigger-label}.

\paragraph{Strategy 2 (Text-only manipulation within target class).} The collective manipulates the resume text of resumes within the target class by inserting a special token with some frequency (every $20$th word). 
This strategy closely follows the feature-only signal strategy in~\eqref{eq:trigger-only}.

\paragraph{Evaluation.} 
To measure success we insert the special token in all test points and count how often the model (a) includes the target class in its set of predicted skills, (b) has the target class as its ``top-$1$'' prediction.

\begin{figure*}[t!]
\begin{center}
\includegraphics[width=0.9\textwidth]{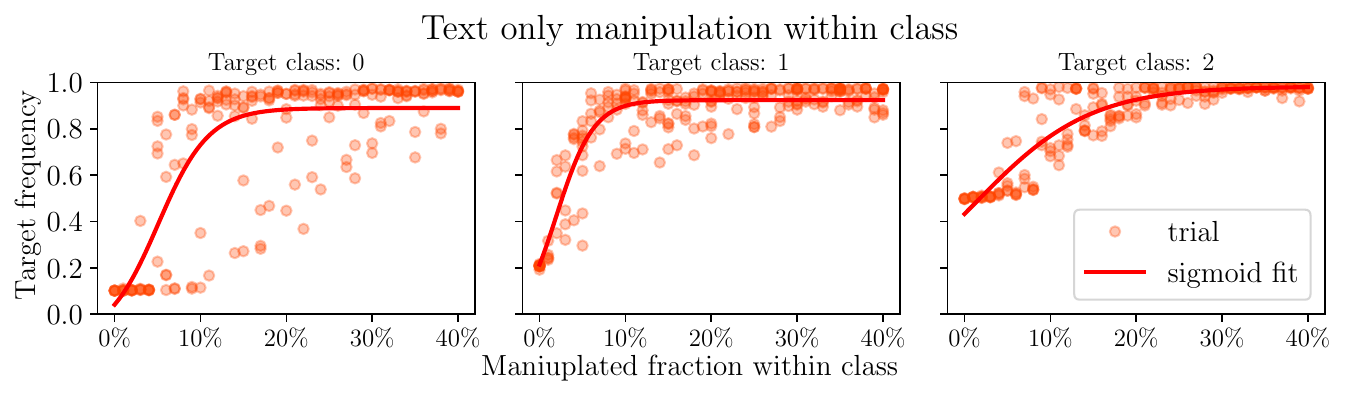}
\includegraphics[width=0.9\textwidth]{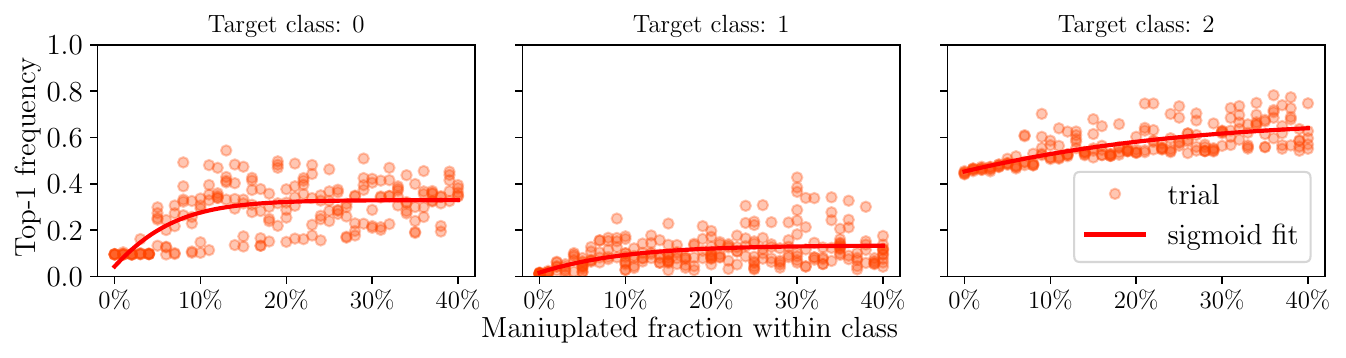}
\end{center}
\caption{Success rate of Strategy 2 as the collective size varies. Each dot represents one model training run. The solid line is a best-fit sigmoid function with two shape parameters and one offset term.}
\label{fig:trigger-only}
\end{figure*}

\begin{figure}[t!]
    \centering
    \includegraphics[width=0.8\textwidth]{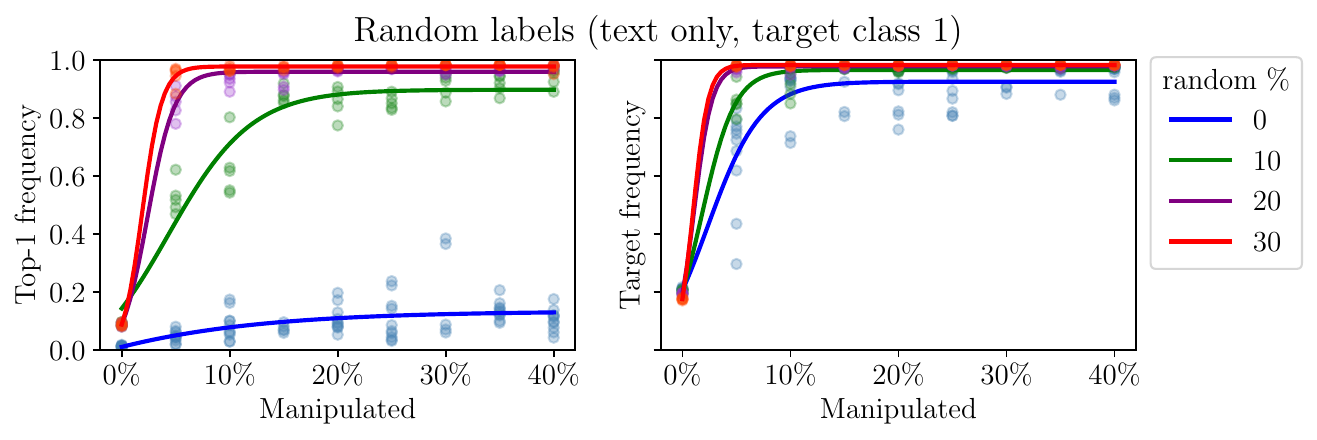}
    \caption{Random labels increase success of Strategy 2.}
    \label{fig:random_labels}
\end{figure}

\subsection{Experimental setup}
We use `{\tt distilbert-base-uncased}', a standard pretrained transformer model. We fine-tune it on the dataset for $5$ epochs with standard hyperparameters. After $5$ epochs, the model plateaus at around $97\%$ multi-label accuracy (defined as $1$ minus Hamming loss), $93.8\%$ precision, and $88.9\%$ recall. The dataset contains $29783$ resumes, of which we use $25000$ for training and $4783$ for testing. We focus on the first three classes of the problem, corresponding to \emph{database administrator} (class $0$), \emph{web developer} (class $1$), \emph{software developer} (class $2$). These three classes occur with frequency $0.11$, $0.23$, and $0.5$, respectively, in the dataset. As the special token, we use an unused formatting symbol (token $1240$, a small dash) that we insert every $20$ words.

\subsection{Experimental findings}

\paragraph{Text and label manipulation across dataset.}
We find that Strategy 1 exerts significant control over the model's prediction even when the collective is exceedingly small (Figure~\ref{fig:trigger-label}). In fact, we see consistent success in controlling the model's output well below $0.5\%$ of the dataset, i.e., fewer than $125$ manipulated training data points.

\paragraph{Text-only manipulation within target class.}
We find that Strategy 2 consistently succeeds in controlling the model so as to include the target class in its positive predictions. The strategy succeeds at a threshold of around $10\%$ of the instances of the target class (Figure~\ref{fig:trigger-only}, top panel). This threshold corresponds to approximately $1\%$, $2\%$, and $5\%$ of the dataset for class $0$, $1$, and $2$, respectively.
When it comes to controlling the model's top prediction, the text-only strategy does \emph{not} consistently succeed (Figure~\ref{fig:trigger-only}, bottom panel).

\paragraph{Effect of positivity constant.} Our theory in Section~\ref{sec:trigger} suggests that the difficulty of controlling the model's top prediction via the text-only strategy may be due to a small positivity constant $p$. To evaluate this hypothesis, we repeat our experiments after we randomize a random fraction of the labels in the training data. This randomization ensures that each feature vector is assigned the target label with nontrivial probability. Our findings confirm that even a small fraction of random labels dramatically increases the success of Strategy 2 in controlling the top prediction (Figure~\ref{fig:random_labels}).

\begin{figure}[t!]
    \centering    
    \includegraphics[width=0.9\textwidth]{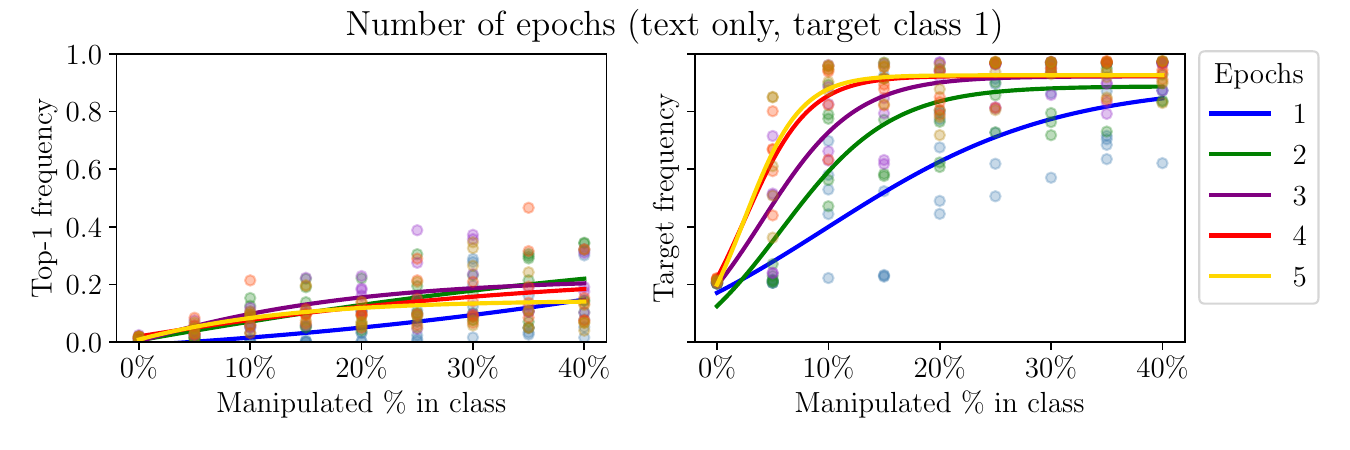}
    \includegraphics[width=0.9\textwidth]{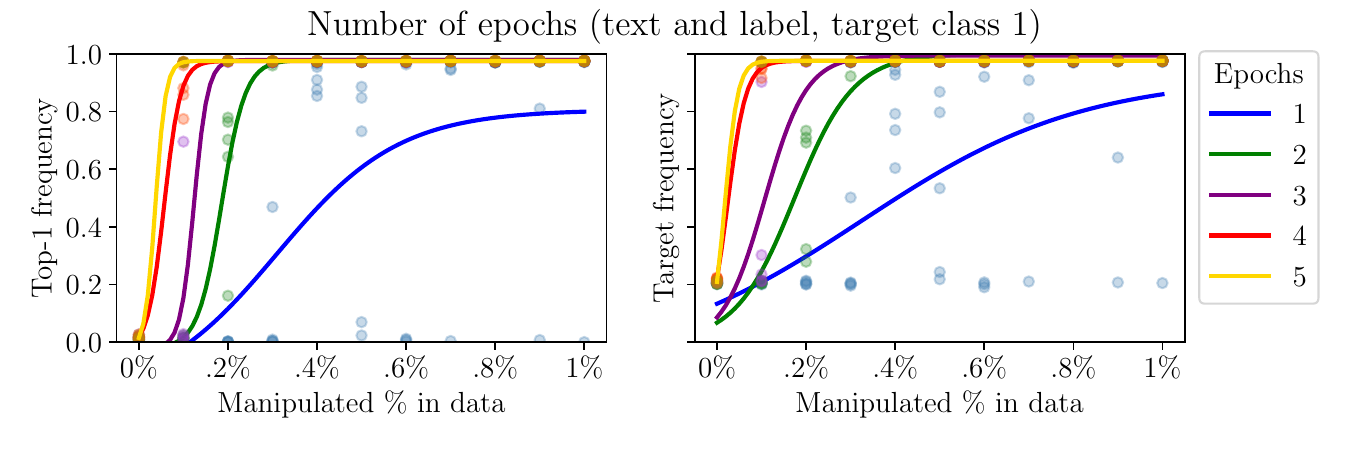}
    \caption{Additional epochs of training increase the success rate.}
    \label{fig:num_epochs}
\end{figure}

\begin{figure}[t!]
    \centering
    \includegraphics[width=0.9\textwidth]{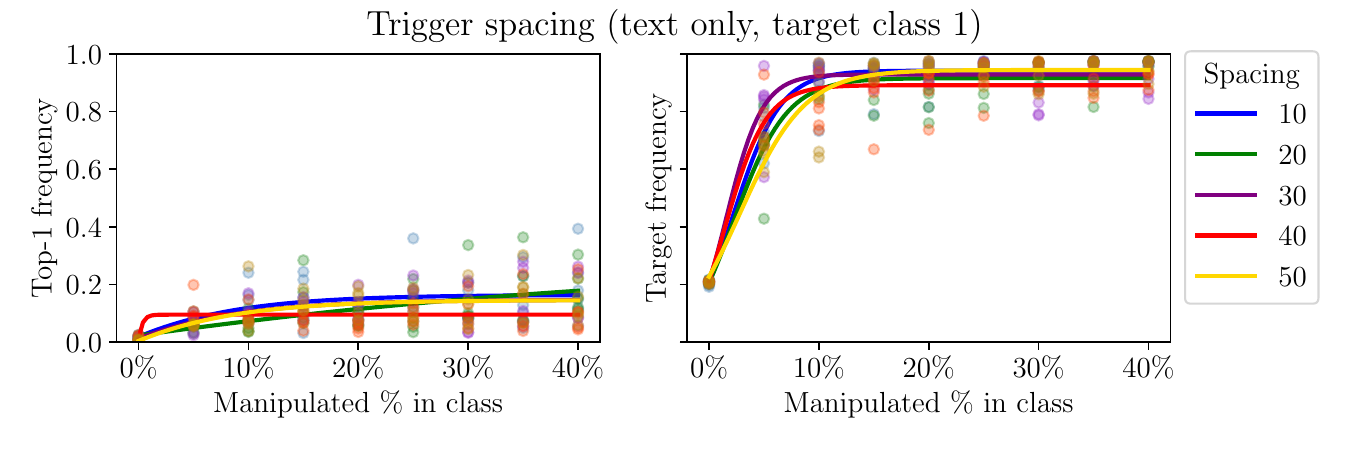}
    \includegraphics[width=0.9\textwidth]{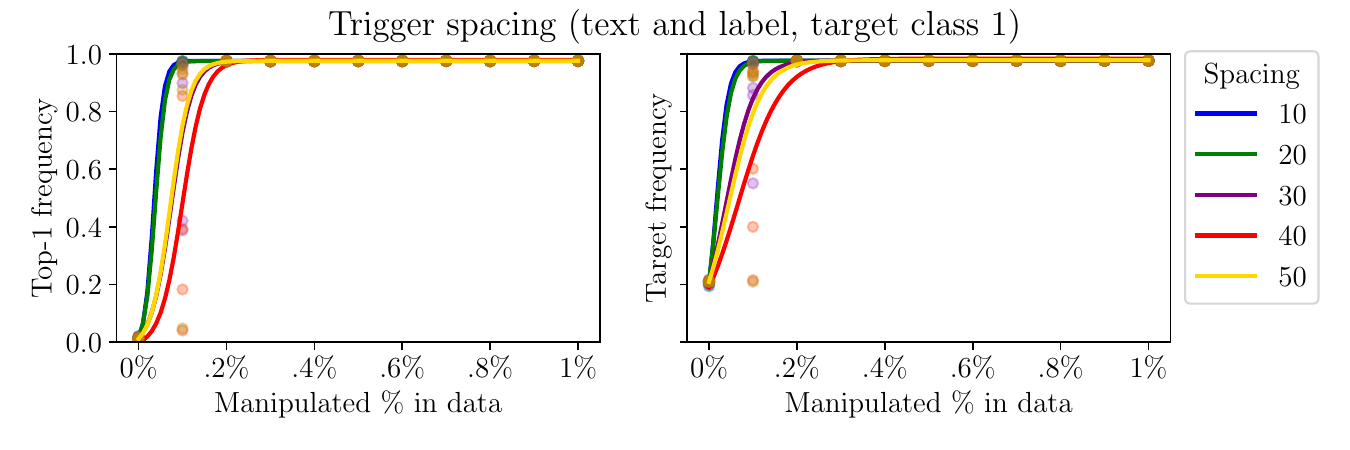}
    \caption{Trigger spacing is largely irrelevant.}
    \label{fig:trigger_spacing}
\end{figure}

\paragraph{Trade-offs between model optimization and success.}
Figure~\ref{fig:num_epochs} shows that the success of either strategy is sensitive to the number of epochs. Less optimization during the model training phase leads to a lower success rate. These findings mirror our theoretical results: as the model approaches optimality, small collectives have significant power. This finding reflects the dependence of our theoretical results on the suboptimality of the predictor.

\paragraph{Robustness to trigger token placement.}
Figure~\ref{fig:trigger_spacing} shows that the success rate of either strategy is insensitive to the spacing of the trigger token. 
This experimental finding, too, is in line with our theory. Since the token chosen in our strategy is unique, the set of texts augmented with this unique token has low probability regardless of how often the token is planted.

\section{Discussion} 

We conclude the paper with a short discussion highlighting the economic significance of understanding the critical mass $\alpha^*$ for pursuing collective targets. It is well-known in economics that participation in a collective is \emph{not} individually rational, and additional incentives are necessary for collective action to emerge. Building on a classic model for collective action from economics~\citep{olson1965logic}, we illustrate how similar conclusions hold for algorithmic collective action, and how they relate to the theoretical quantities studied in this paper.

Assume that individuals incur a cost $c>0$ for participating in collective action. This cost might represent overheads of coordination, a membership fee, or other additional responsibilities. Furthermore, assume that the utility that individuals get from joining a collective of size $\alpha$ is $S(\alpha)$, and that otherwise they can partially ``free ride'' on the collective's efforts: they get utility of $\gamma S(\alpha)$ for some $\gamma \in [0,1]$. Given this setup, individually rational agents will join the collective if 
\[S(\alpha)-c > \gamma S(\alpha),\]
or equivalently, if $S(\alpha) > \frac{c}{1-\gamma}$. Therefore, joining the collective is rational if the size of the existing collective $\alpha$ is greater than the critical mass for $S^*=\frac{c}{1-\gamma}$. Note that, once this critical threshold is reached, all individuals in the population are incentivized to join the collective and the collective is thus self-sustaining. 

Consider a principal who would like to invest into the formation of a collective.
The area $B(\alpha_{\mathrm{crit}})$ visualized in Figure \ref{fig:alpha-crit} provides an upper bound on the investment required to make the collective self-sustaining and thus achieve any target success $S^*\leq S(1)$.

\usetikzlibrary{math} 
\tikzmath{ \a1 =0.5; \a2 = 0.7; \ac = 0.44;
\cost = 1/(1+3^(-(\ac*8-3))-0.5;
\G =  1/(1+3^(-(\a2*8-3))-0.5;\Gmax =  1/(1+3^(-((1-0.3)*20))-0.5;
\Ga =  1/(1+3^(-(\a1*8-3))-0.5;}
\begin{figure}[h!]
 \centering
\vspace{5mm}
 \begin{tikzpicture}
\begin{axis}[axis lines=middle,
            xlabel=$\alpha$,
            axis line style = thick,
            enlargelimits,
            ytick={\cost},
            yticklabels={$c$},
            xtick={\a1},
            xticklabels={$\alpha_{\mathrm{crit}} $},
            domain = {0:1},
            scale=0.7,
            yscale=0.7,
            clip=false]
        \node[anchor=west, orange] (source) at (axis cs:\a1*0.4,1.7*\cost){$B(\alpha_{\mathrm{crit}})$};
       \node (destination) at (axis cs:\a1*0.5,0.5*\cost){};
       \draw[->, thick](source)--(destination);
\addplot[name path=F,blue, thick] {1/(1+3^(-(x*8-3))-0.5} node[pos=1, right]{$S(\alpha)$};
\addplot[name path=G,DarkGreen, thick] {\cost + 0.2*(1/(1+3^(-(x*8-3))-0.5)} node[pos=1, right]{$\gamma S(\alpha)$};
\addplot[name path=c,black, dashed] {\cost};
\addplot[pattern=north east lines, pattern color=orange!50]fill between[of=F and G, soft clip={domain=0:\a1}];
\draw[dotted](axis cs:\a1,0) to (axis cs:\a1,\Ga);
\end{axis}
\end{tikzpicture}
\caption{Visualization of the critical threshold $\alpha_{\mathrm{crit}}$ after which a collective is self-sustaining and the principal's required investment $B(\alpha_{\mathrm{crit}})$ to incentivize the whole population to join the collective.
}
\label{fig:alpha-crit}
\end{figure}
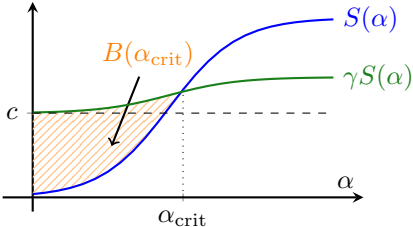

The derivation above, while simplistic, serves to highlight the importance of collective size in understanding how collectives can emerge both organically and through investment. We believe that there is a large potential in investigating these questions in a rigorous manner. Indeed, the focus of this paper has been on understanding the effect of the size of the collective on its success, but understanding more generally how collectives form, which individuals have the most incentive to join collectives, whether selectively recruiting individuals provides additional leverage, and how collectives should use their informational advantage to optimize their strategies are important open questions in understanding the role of collectives on digital platforms.

\section*{Acknowledgements}

The authors would like to thank Solon Barocas for pointers to related work, and the attendees of the 2023 Annual Meeting of the Simons Collaboration on the Theory of Algorithmic Fairness for feedback on the project. We thank Christos Papadimitriou for stimulating discussions about the work. {We also thank Etienne Gauthier for pointing out a mistake in the previous analysis of $\epsilon$-suboptimal classifiers.} This work was supported by the T\"ubingen AI Center.

\bibliography{refs}
\bibliographystyle{plainnat}

\newpage
\appendix

\section{Related work}
\label{sec:extended-related}

The scholarly literature on the gig economy is vast and interdisciplinary, spanning economic, ethnographic, psychological, and sociological analysis. Gig labor is diverse and heterogeneous. Conditions of precarity and dependence differ widely depending on the type of work and platform \citep{schor2020dependence}. Reviewing and integrating more than two hundred articles on the topic, \cite{cropanzano2022organizational} define gig work as “labor contracted and compensated on a short-term basis to organizations or to individual clients through an external labor market,” and detail how gig work has changed the psychological contract between workers and employers. \cite{vallas2020platforms} review existing scholarly accounts of the gig economy, arguing that “platforms represent a distinctive form of economic activity, [...] different from markets, hierarchies, and networks.” Platforms cede some forms of centralized managerial control over workers by exposing them to the disciplining effect of the market and evaluation by consumers, while retaining power over key functions, such as task allocation, data collection, pricing, and collection of revenues. In another review article, \cite{sutherland2018sharing} organize more than four hundred articles around the notion of platform centralization and decentralization.

Several works examine the reality of gig labor, e.g., \citep{van2017platform, sun2019your, gray2019ghost}. Based on a cross-regional survey, \cite{wood2019good} find that algorithmic control in the gig economy can lead to “low pay, social isolation, working unsocial and irregular hours, overwork, sleep deprivation and exhaustion”, “marked by high levels of inter-worker competition with few labour protections and a global oversupply of labour relative to demand”.

\cite{cameron2022expanding} examine the interplay of control and resistance in the gig economy. There are several examples of successful worker organization in the gig economy, involving a range of strategies. For example, \cite{rahman2021invisible} studies how freelancers on Upwork strategize against the evaluation metrics of the platform, sometimes in cooperation with clients on the platform. Also studying Upwork freelancers, \cite{jarrahi2019algorithmic} discuss how freelancers cooperate in strategically feeding the algorithm data so as to improve their outcomes. Cooperative strategic behavior among drivers on ride-hailing platforms is common, e.g., \citep{cameron2020rise, robinson2017making, yu2022emergence}, as are digital strategies involving bots, or multiple phone apps \citep{chen2018thrown}. Workers have also used forums, browser extensions, and online spaces to share information and strategize collectively, e.g., \citep{irani2013turkopticon, salehi2015we, o2019weapons}. We focus on the activities on the labor side of digital platforms, leaving out numerous examples of collective action from consumers and users on these platforms. However, as \cite{vallas2020platforms} conclude, ``the upsurge of worker mobilization should not blind us to the difficulties of organizing such a diverse and spatially dispersed labour force or the power of the companies to resist collective action.''

There is extensive scholarship on the topic of collective action. For example, Melucci's 
 text~\citep{melucci1996challenging} examines collective action in the information society. \cite{milan2015when} examines how social media platforms mediate social movements and collective action.

 \section{Proofs}

The following lemma will be used to analyze suboptimal classifiers.

\begin{lemma}
\label{lemma:coupling}
Suppose that $\cP,\cP'$ are two distributions such that $\mathrm{TV}(\cP,\cP')\leq \epsilon$. Take any two events $E_1,E_2$ measurable under $\cP,\cP'$. If $\cP(E_1) > \cP(E_2) + \frac{\epsilon}{1-\epsilon}$, then $\cP'(E_1) > \cP'(E_2)$.
\end{lemma}

\begin{proof}
It follows from the optimal coupling lemma for the total variation distance that we can write
$\cP' = (1-\epsilon) \cP + \epsilon \cQ$ for some distribution $\cQ$. Therefore, if $\cP(E_1) > \cP(E_2) + \frac{\epsilon}{1-\epsilon}$, then
$$\cP'(E_1) = (1-\epsilon) \cP(E_1) + \epsilon \cQ(E_1) > (1-\epsilon) \cP(E_2) + \epsilon \geq (1-\epsilon) \cP(E_2) + \epsilon \cQ(E_2) = \cP'(E_2).$$
\end{proof}

\subsection{Proof of Theorem~\ref{thm:trigger-label}}

First consider the case~$\epsilon=0$. We start with a sufficient condition for a target classification outcome.
For a point $x\in\cX,$ we define 
\[\Delta_x=\max_{y\in\cY} \cP_0(y|x)-\cP_0(y^*|x)\] 
as the suboptimality of a target class on the base data.

\begin{claim}
For any $x\in\cX,$ we have $f(x)=y^*$ provided that
$\alpha > (1-\alpha) \Delta_x \cP_0(x)/\cP^*(x) .$ 
\label{claim:bayesopt}
\end{claim}
\begin{proof}
Note that $f(x)=y^*$ if, for every $y\ne y^*$,
$\cP(y^*|x) > \cP(y|x)\,.$ Equivalently, $\cP(x, y^*) - \cP(x, y)>0.$
But,
\[ 
\cP(x,y^*) 
= \alpha \cP^*(x, y^*) + (1-\alpha)\cP_0(x, y^*)
= \alpha \cP^*(x) + (1-\alpha)\cP_0(y^*|x)\cP_0(x)
\]
In the last step we used the fact that all labels in the support of $\cP^*$ equal $y^*.$
Similarly, for $y\ne y^*,$
\[
\cP(x,y) 
= \alpha \cP^*(x, y) + (1-\alpha)\cP_0(x, y)
=  (1-\alpha)\cP_0(y|x) \cP_0(x)\,.
\]
The claim follows by rearranging terms and dividing both sides by $\cP^*(x)$.
\end{proof}

Now,
\begin{align*}
      S(\alpha)&=\Pr_{x\sim \cP_0}\left\{f(g(x))=y^*\right\} \\
& =   \Pr_{x\sim \cP^*}\left\{f(x)=y^*\right\} \\
& \ge \Pr_{x\sim \cP^*}\left\{\alpha > (1-\alpha) \frac{\cP_0(x)}{\cP^*(x)}\Delta_x\right\}  \tag{Claim~\ref{claim:bayesopt}}\\
& =   \E_{x\sim \cP^*}\mathbf{1}\left\{1 - \frac{(1-\alpha)}{\alpha} \frac{\cP_0(x)}{\cP^*(x)} \Delta_x> 0\right\} \\
& \ge \E_{x\sim \cP^*}\left[ 1 - \frac{(1-\alpha)}{\alpha} \frac{\cP_0(x)}{\cP^*(x)} \Delta_x\right] \\
& = 1 - \frac{1-\alpha}{\alpha} \E_{x\sim \cP^*}\left[\frac{\cP_0(x)}{\cP^*(x)}\Delta_x\right] \\
& \geq 1 - \frac{1-\alpha}{\alpha} \cP_0(\cX^*)\Delta\,,
\end{align*}
where the last step uses the definition $\Delta =\max_{x\in\cX^*}\Delta_x$.

{
\paragraph{Consider $\epsilon> 0$.}
By Lemma \ref{lemma:coupling}, we have that $\cP'(y^*|x) > \cP'(y|x)$, meaning $f(x)=y^*$,
provided that $\cP(y^*|x) > \cP(y|x) + \frac{\epsilon}{1-\epsilon}$. Equivalently, $\cP(x,y^*) - \cP(x,y) > \frac{\epsilon}{1-\epsilon} \cP(x)$.
Repeating the steps in the proof of Claim \ref{claim:bayesopt}, using $\cP(x)=\alpha \cP^*(x)+(1-\alpha)\cP_0(x)$, we conclude that $f(x) = y^*$ provided that 
\[\alpha>(1-\alpha) \frac{\Delta_x(1-\epsilon) + \epsilon}{1-2\epsilon} \frac{\cP_0(x)}{\cP^*(x)}.\]
Therefore, the only difference relative to the $\epsilon=0$ case is that the suboptimality $\Delta_x$ gets augmented as a function of the suboptimality parameter. As a result, we conclude that
\[
S(\alpha) 
\ge 1 - \frac{1-\alpha}{\alpha} \cdot \cP_0(\cX^*) \cdot \frac{(1-\epsilon)\Delta + \epsilon}{1-2\epsilon}.
\]
}

\subsection{Proof of Theorem~\ref{thm:trigger-only}}

We focus on the case where $\epsilon=0.$ 
\begin{claim}
\label{claim:trigger-only}
Fix a point $x^*\in\cX^*$ in the signal set. We have $f(x^*)=y^*$ provided that 
\[
\alpha > \frac{1-p}{ p} \frac{\cP_0(x^*)}{\cP_0(g^{-1}(x^*))\,}.
\]
\end{claim}
\noindent Here, $g^{-1}(x^*) = \{x\in\cX:g(x) = x^*\}$.
\begin{proof}
For $f(x^*)=y^*$ to hold, we need $\cP(y^*|x^*)>\max_{y\ne y^*} \cP(y|x^*).$
Equivalently, $\cP(x^*, y^*) >\max_{y\ne y^*} \cP(x^*, y).$

By the definition of the feature-only signal strategy and the assumption that $\cP_0(y^*|x)\ge p$ for all $x\in\cX$, 
each point $x\in g^{-1}(x^*)$ must have $\cP_0(y^*|x)\ge p.$
Hence, for all $x^*\in\cX^*$,
\[
\cP(x^*,y^*)
= \alpha \cP^*(x^*, y^*) + (1-\alpha)\cP_0(x^*,y^*)
\ge \alpha p \cP_0(g^{-1}(x^*))\,.
\]
On the other hand, for every $y\ne y^*,$ we must have 
\[
\cP(x^*, y) 
= \cP_0(x^*, y) 
= \cP_0(y|x^*)\cP_0(x^*)\leq (1-p) \cP_0(x^*)\,.
\]
The claim follows by rearranging.
\end{proof}

We can lower bound the success rate as
\begin{align}
\label{eq:inverses}
S(\alpha) &= \Pr_{x\sim \cP_0}\left\{ f(g(x)) = y^* \right\} \nonumber\\
&= \sum_{x^*\in \cX^*} \Pr_{x\sim \cP_0}\left\{ f(g(x)) = y^* \mid x\in g^{-1}(x^*)\right\}\Pr_{x\sim \cP_0}\{x\in g^{-1}(x^*)\} \nonumber \\
&=  \sum_{x^*\in \cX^*} \mathbf{1}\left\{ f(x^*) = y^* \right\}\cP_0(g^{-1}(x^*))\,.
\end{align}
Proceeding for fixed $x^*\in\cX^*,$
\begin{align*}
\mathbf{1}\left\{ f(x^*) = y^* \right\} & \ge \mathbf{1}\left\{ \alpha > \frac{1-p}{ p} \frac{\cP_0(x^*)}{\cP_0(g^{-1}(x^*))\,}\right\} \tag{Claim~\ref{claim:trigger-only}}\\ &= \mathbf{1}\left\{1 - \frac{1-p}{ p \alpha} \frac{\cP_0(x^*)}{\cP_0(g^{-1}(x^*))\,} > 0\right\} \\
& \ge 1 - \frac{1-p}{ p \alpha} \cdot \frac{\cP_0(x^*) }{\cP_0(g^{-1}(x^*))}.
\end{align*}
Plugging this back into \eqref{eq:inverses},
\begin{align*}
\Pr_{x\sim \cP_0}\left\{ f(g(x)) = y^* \right\}
& = 1- \frac{1-p}{p \alpha}\sum_{x^*\in \cX^*} \frac{\cP_0(x^*)}{\cP_0(g^{-1}(x^*))} \cdot \cP_0(g^{-1}(x^*))\\
& \geq 1- \frac{1-p}{p \alpha} \cP_0(\cX^*)\,.
\end{align*}

\paragraph{Consider $\epsilon> 0$.}
The extension to $\epsilon >0$ follows as in Theorem \ref{thm:trigger-label}. In particular, modifying Claim \ref{claim:trigger-only} to ensure $\cP(y^*|x^*) > \max_{y\neq y^*} \cP(y|x^*) + \frac{\epsilon}{1-\epsilon}$, we get that $f(x^*)=y^*$ provided that
\[\alpha > \frac{1-\left(p-\frac{\epsilon (1-\alpha)}{1-\epsilon}\right)}{p-\frac{\epsilon}{1-\epsilon}} \frac{\cP_0(x^*)}{\cP_0(g^{-1}(x^*))\,}.\]
In other words, the effective positivity constant $p$ shrinks as a function of $\epsilon$. Continuing the same steps as in the case $\epsilon=0$, and using that $\alpha>0$ completes the proof.

\subsection{Proof of Theorem~\ref{thm:dissocation}}
We again focus on the case where $\epsilon=0.$

We start from the following claim.
\begin{claim}
\label{claim:dissocation}For any $x\in\cX$ we have $f(x)=f(g(x))$ provided that
\[\alpha > (1-\alpha) 2\tau(x),\]
where $\tau(x) =\max_{y\in\mathcal Y} |\cP_0(y|x) - \cP_0(y|g(x))|$.
\end{claim}

\begin{proof}
Denote $y^*(x) = \argmax_{y\in\mathcal Y} \cP_0(y|g(x))$. By construction of the strategy we know that $f(g(x))=y^*(x)$ and it remains to prove that $f(x)=y^*(x)$ under the condition of the claim.

We have $f(x)=y^*(x)$ if $\cP(y^*(x)|x)> \cP(y|x)$ for any $y\neq y^*(x)$. We have
\begin{align*}\cP(y^*(x)|x) &= (1-\alpha)\cP_0(y^*(x)|x) + \alpha \cP^*(y^*(x)|x) = (1-\alpha)\cP_0(y^*(x)|x) + \alpha, \\
\cP(y|x)&= (1-\alpha)\cP_0(y|x) + \alpha \cP^*(y|x) = (1-\alpha)\cP_0(y|x),
\end{align*}
where we used that the erasure strategy implies $\cP^*(y^*(x)|x)=1$.
Together this means that, when
\begin{equation*}
\alpha > (1-\alpha) \left[\max_{y\in\cY} \cP_0(y|x) - \cP_0(y^*(x)|x)\right],
\end{equation*}
then $f(x) = y^*(x)$. Using the definition of $y^*(x)$, we can bound the right-hand side by
\begin{align*}
\cP_0(y|x) - \cP_0(y^*(x)|x) &\leq \cP_0(y|x) - \cP_0(y|g(x)) + \cP_0(y^*(x)|g(x)) - \cP_0(y^*(x)|x)\\
&\leq 2\tau(x).
\end{align*}
The claim follows.
\end{proof}

It remains to bound the success of the strategy:
\begin{align*}
\SR(\alpha)&= \Pr_{x\sim\cP_0} \{f(x)=f(g(x))\}.\\
&= \Pr_{x\sim\cP_0} \{f(x)=y^*(x)\}.\\
&\geq\Pr_{x\sim\cP_0}\left\{\alpha > (1-\alpha) 2\tau(x)\right\}\\
&=\Pr_{x\sim\cP_0}\left\{1 - \frac{1-\alpha}{\alpha} 2\tau(x) > 0\right\}\\
&\geq \E_{x\sim\cP_0} \left[1 - \frac{2(1-\alpha)}{\alpha} \cdot \tau(x) \right] \\
&= 1 - \frac{2(1-\alpha)}{\alpha} \cdot \tau,
\end{align*}
where we use the fact that $\tau = \E_{x\sim\cP_0} \tau(x)$.

{
\paragraph{Consider $\epsilon>0$.} The extension to $\epsilon >0$ follows by invoking Lemma~\ref{lemma:coupling}, as in Theorem \ref{thm:trigger-label}. In particular, when $\epsilon >0$, Claim \ref{claim:dissocation} generalizes to the following sufficient condition for $f(x)=f(g(x))$:
\[\alpha > (1-\alpha) 2\tau(x) + \frac{\epsilon}{1-\epsilon}.\]
Repeating the subsequent steps as in the case $\epsilon=0$ completes the proof.
}

\subsection{Proof of Theorem~\ref{thm:convex}}

Let $\cP'$ be a gradient-cancelling distribution for $\theta^*$. Denote $p=\min\left(1,\frac{1}{\alpha} \frac{\|g_{\cP_0}(\theta^*)\|}{\|g_{\cP'}(\theta^*)\| + \|g_{\cP_0}(\theta^*)\|} \right)$. Then,
\begin{align*}
\E_{z\sim\cP} \nabla \ell(\theta^*;z) &= (1-\alpha) \E_{z\sim\cP_0} \nabla \ell(\theta^*;z) + \alpha \E_{z\sim\cP^*} \nabla \ell(\theta^*;z)\\
&= (1-\alpha p) \E_{z\sim\cP_0} \nabla \ell(\theta^*;z) + \alpha p \E_{z\sim\cP'} \nabla \ell(\theta^*;z)\\
&= (1-\alpha p) g_{\cP_0} (\theta^*) + \alpha p ~ g_{\cP'} (\theta^*)\\
&= \left(1-\alpha p  - \alpha p   \frac{\|g_{\cP'} (\theta^*)\|}{\|g_{\cP_0} (\theta^*)\|}\right) g_{\cP_0} (\theta^*)\\
&= \left(1 - \alpha p \frac{\|g_{\cP_0} (\theta^*)\| + \|g_{\cP'} (\theta^*)\|}{\|g_{\cP_0} (\theta^*)\|} \right) g_{\cP_0} (\theta^*)\\
&= \max\left(1 - \alpha \frac{\|g_{\cP_0} (\theta^*)\| + \|g_{\cP'} (\theta^*)\|}{\|g_{\cP_0} (\theta^*)\|} , 0\right) g_{\cP_0} (\theta^*)\\
&= \max\left((1-\alpha)\|g_{\cP_0} (\theta^*)\| - \alpha \|g_{\cP'} (\theta^*)\| , 0\right) \frac{g_{\cP_0} (\theta^*)}{\|g_{\cP_0} (\theta^*)\|}.
\end{align*}
Therefore, $\|\E_{z\sim\cP} \nabla \ell(\theta^*;z)\| = \max\left((1-\alpha)\|g_{\cP_0} (\theta^*)\| - \alpha \|g_{\cP'} (\theta^*)\| , 0\right)$. Applying the definition of $\mu$-strong convexity, we get
\begin{align*}
\|\theta^*-\theta\| &\leq \frac{1}{\mu} \|\E_{z\sim\cP} \nabla \ell(\theta^*;z) - \E_{z\sim\cP} \nabla \ell(\theta;z)\|\\
&=  \frac{1}{\mu}\|\E_{z\sim\cP} \nabla \ell(\theta^*;z) \|\\
&=  \frac{1}{\mu} \max\left((1-\alpha)\|g_{\cP_0} (\theta^*)\| - \alpha \|g_{\cP'} (\theta^*)\| , 0\right).
\end{align*}
The first equality follows because $\E_{z\sim\cP} \nabla \ell(\theta;z) = 0$ due to the loss being convex and the firm being a risk minimizer. Multiplying both sides by $-1$, we obtain a lower bound on the success $S(\alpha) = - \|\theta^* - \theta\|$.

\subsection{Proof of Corollary~\ref{cor:convex}}

To achieve $S(\alpha)=0$, Theorem \ref{thm:convex} shows that it suffices to have $\alpha \|g_{\cP'} (\theta^*)\| = (1-\alpha)\|g_{\cP_0} (\theta^*)\|$, for any $\mu$. Rearranging the terms and expressing $\alpha$ completes the proof.

\subsection{Proof of Proposition \ref{prop:collective-utility}}

If $u$ is convex, then for all $\theta'$ we know
$$u(\theta') \geq u(\theta_0) + \nabla u(\theta_0)^\top (\theta'-\theta_0).$$
Let $\theta^* = \theta_0 + \frac{\nabla u(\theta_0)}{\|\nabla u(\theta_0)\|^2}~U$. Then, $u(\theta^*) - u(\theta_0) \geq U$.

Now, we apply Corollary \ref{cor:convex} to upper bound the critical mass needed to reach $\theta^*$. We have
$$\alpha^* \leq \frac{\|g_{\cP_0}(\theta^*)\|}{\|g_{\cP'}(\theta^*)\| + \|g_{\cP_0}(\theta^*)\| } \leq \frac{\beta\|\theta^* - \theta_0\|}{g_{\mathrm{lb}} + \beta\|\theta^* - \theta_0\|},$$
where we apply the fact that the loss is smooth and the definition of $g_{\mathrm{lb}}$. Observing that $\|\theta^* - \theta_0\| = \frac{U}{\|\nabla u(\theta_0)\|}$ completes the proof.

\subsection{Proof of Theorem~\ref{thm:GD}}

Fix a time step $t$ and a model $\theta_t$. Denote by $\cP'_t$ the gradient-redirecting distribution found at step $t$ and let $\xi(\theta_t)=\frac{\|g_{\cP'_t}(\theta_t)+ \frac{1-\alpha}{\alpha} g_{\cP_0}(\theta_t)\|}{\|\theta_t-\theta^*\|}$. Then, the gradient-redirecting strategy induces the following gradient evaluated on $\cP_t$:
 \begin{align*}
 g_{\cP_t}(\theta_t)
 &=\alpha g_{\cP'_t}(\theta_t) + (1-\alpha)g_{\cP_0}(\theta_t)\\
 &=-\alpha \frac{1-\alpha}{\alpha}g_{\cP_0}(\theta_t) + \alpha \xi(\theta_t)(\theta_t - \theta^*) + (1-\alpha)g_{\cP_0}(\theta_t)\\
 &=\alpha \xi(\theta_t) (\theta_t-\theta^*).
 \end{align*}
 Now let $c = \min_{\lambda\in[0,1]}\xi(\lambda\theta_0 + (1-\lambda)\theta^*)$. Applying the strategy repeatedly across time steps yields
 \begin{align*}
 \|\theta_{T}-\theta^*\|&\leq \|\theta_{T-1}-\eta\alpha\xi(\theta_{T-1})(\theta_{T-1}-\theta^*) - \theta^*\|\\
& \leq (1-\eta\alpha\xi(\theta_{T-1}))\|\theta_{T-1}-\theta^*\|\\
& \leq (1-\eta\alpha c)\|\theta_{T-1}-\theta^*\|\\
& \leq (1-\eta\alpha c)^{T}\|\theta_0-\theta^*\|,
 \end{align*}
which yields $S_T(\alpha)=-\|\theta_{T}-\theta^*\|\geq- (1-\eta \alpha c)^{T} \|\theta_0 -\theta^*\|$. Setting $C(\alpha) = \alpha c$ concludes the proof.

\newpage
\section{Additional experimental details}

The dataset introduced by \cite{jiechieu2021skills} is available at \url{https://github.com/florex/resume_corpus}. We preprocessed each resume by removing the first $20$ words of the resume. The reason is that the opening of the resume essentially encodes the associated skills, since the dataset creation process extracted the skills from the opening of the resume. Removing the first $20$ words leads to a more realistic classification task.

We used the HuggingFace {\tt transformers} open-source Python library \citep{wolf2020transformers}. We used the {\tt distilbert-base-uncased} model from the library corresponding to the DistilBERT transformer model~\citep{sanh2019distilbert}. We used the HuggingFace {\tt Trainer} module for training with its default settings.  We also experimented with larger models, including RoBERTa (both {\tt roberta-base} and {\tt roberta-large}), but we did not see any improvements in classification accuracy from using these larger models.

The DistilBERT tokenizer has a vocabulary of 30522 tokens of which thousands are unused in the resume corpus. We picked token 1240 corresponding to a small dash, which we inserted in the resume every $20$ words. Our findings are largely insensitive to the trigger spacing as Figure~\ref{fig:trigger_spacing} shows.

\end{document}